%% file: main.tex
\documentclass[10pt,twocolumn,letterpaper]{article}

 \usepackage{cvpr}              

\input{preamble}

\definecolor{cvprblue}{rgb}{0.21,0.49,0.74}
\usepackage[pagebackref,breaklinks,colorlinks,citecolor=cvprblue]{hyperref}
\usepackage{multirow}
\usepackage{array}
\usepackage{booktabs}
\usepackage{bbding}
\usepackage[accsupp]{axessibility} 

\usepackage{color}

\def\paperID{3508} 
\def\confName{CVPR}
\def\confYear{2024}

\title{SDPose: Tokenized Pose Estimation via Circulation-Guide Self-Distillation}

\author{
    \quad Sichen Chen$^{1}$\thanks{Equal Contribution.}
    \quad Yingyi Zhang$^{2*}$
    \quad Siming Huang$^{2*}$
    \quad Ran Yi$^{1}$
    \quad Ke Fan$^{1}$
    \\
    \quad Ruixin Zhang$^{2}$
    \quad Peixian Chen$^{2}$
    \quad Jun Wang$^{3}$
    \quad Shouhong Ding$^{2}$\thanks{Corresponding Authors.}
    \quad Lizhuang Ma$^{1,4\dagger}$
    \vspace{1em}
    \\
    $^{1}$Shanghai Jiao Tong University, 
    $^{2}$Tencent Youtu Lab, 
    $^{3}$Tencent WeChat Pay Lab33
    \\
    $^{4}$MoE Key Lab of Artificial Intelligence, Shanghai Jiao Tong University
    }
    
\begin{document}
\maketitle

\begin{abstract}
    Recently, transformer-based methods have achieved state-of-the-art prediction quality on human pose estimation(HPE).
    Nonetheless, most of these top-performing transformer-based models are too computation-consuming and storage-demanding to deploy on edge computing platforms.
    Those transformer-based models that require fewer resources are prone to under-fitting due to their smaller scale and thus perform notably worse than their larger counterparts.
    Given this conundrum, we introduce SDPose, a new self-distillation method for improving the performance of small transformer-based models. 
    To mitigate the problem of under-fitting, we design a transformer module named Multi-Cycled Transformer(MCT) based on multiple-cycled forwards to more fully exploit the potential of small model parameters.
    Further, in order to prevent the additional inference compute-consuming brought by MCT, we introduce a self-distillation scheme,
    extracting the knowledge from the MCT module to a naive forward model.
    Specifically, on the MSCOCO validation dataset, SDPose-T obtains 69.7\% mAP with 4.4M parameters and 1.8 GFLOPs. 
    Furthermore, SDPose-S-V2 obtains 73.5\% mAP on the MSCOCO validation dataset with 6.2M parameters and 4.7 GFLOPs, achieving a new state-of-the-art among predominant tiny neural network methods.
    Our code is available at https://github.com/MartyrPenink/SDPose. 
\end{abstract} 

\begin{figure}[t]
    \centering
    \includegraphics[width=1\linewidth]{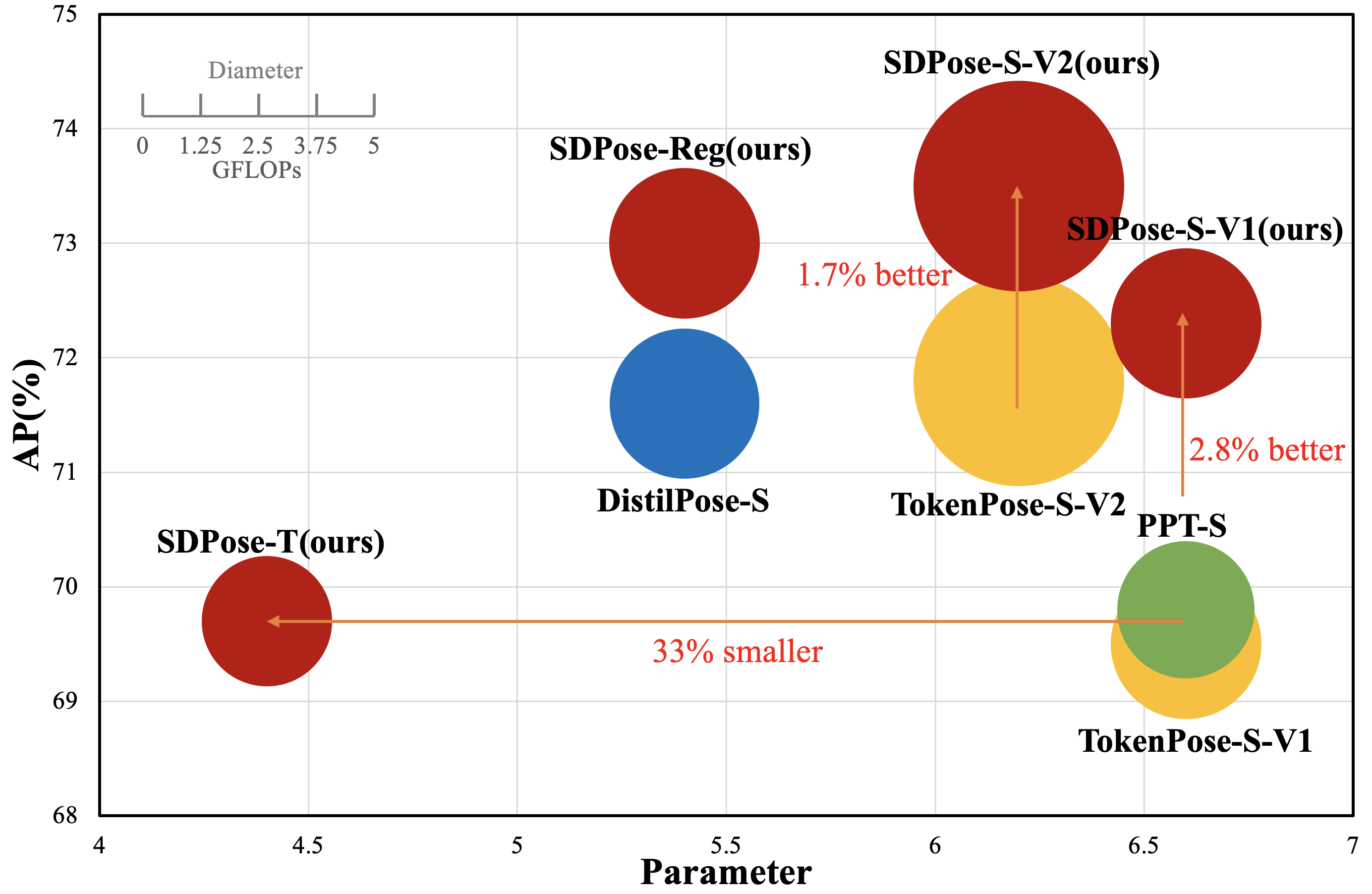}
    \caption[width=0.95\linewidth]{Comparsions between other small models and our methods on MSCOCO validation dataset. Compared to other methods, our approach can significantly reduce the scale while maintaining the same performance, or greatly improve performance under the same scale.}
    \label{fig:parameter_ap}
    \vspace{-0.6cm}
\end{figure}

\vspace{-0.4cm}

\section{Introduction}
    Human Pose Estimation (HPE) aims to estimate the position of each joint point of the human body in a given image. 
    HPE tasks support a wide range of downstream tasks such as activity recognition\cite{driveractivity}, motion capture\cite{motion_capture_survey}, etc.
    Recently with the ViT model being proven effective on many visual tasks, 
    many transformer-based methods\cite{Tokenpose,transpose,vitpose} have achieved excellent performance on HPE tasks. 
    %
    Compared with past CNN-based methods\cite{hrnet}, transformer-based models are much more powerful in capturing the relationship between visual elements.
    However, most of them are large and computationally expensive.
    The state-of-the-art(SOTA) transformer-based model\cite{vitpose} has 632 million parameters and requires 122.9 billion floating-point operations.
    Such a large-scale model is difficult to deploy on edge computing devices and cannot accommodate the growing development of embodied intelligence.
    %
    However, when the CNN or ViT used as a backbone is not of sufficient scale, 
    transformer-based models are not able to learn the relationship between keypoints and visual elements well resulting in poor performance.
    Stacking more transformer layers is a viable approach\cite{transpose}, but this also increases the scale of the network resulting in larger parameters and the difficulty of edge deployment.

\begin{figure}[t]
    \centering
    \includegraphics[width=1\linewidth]{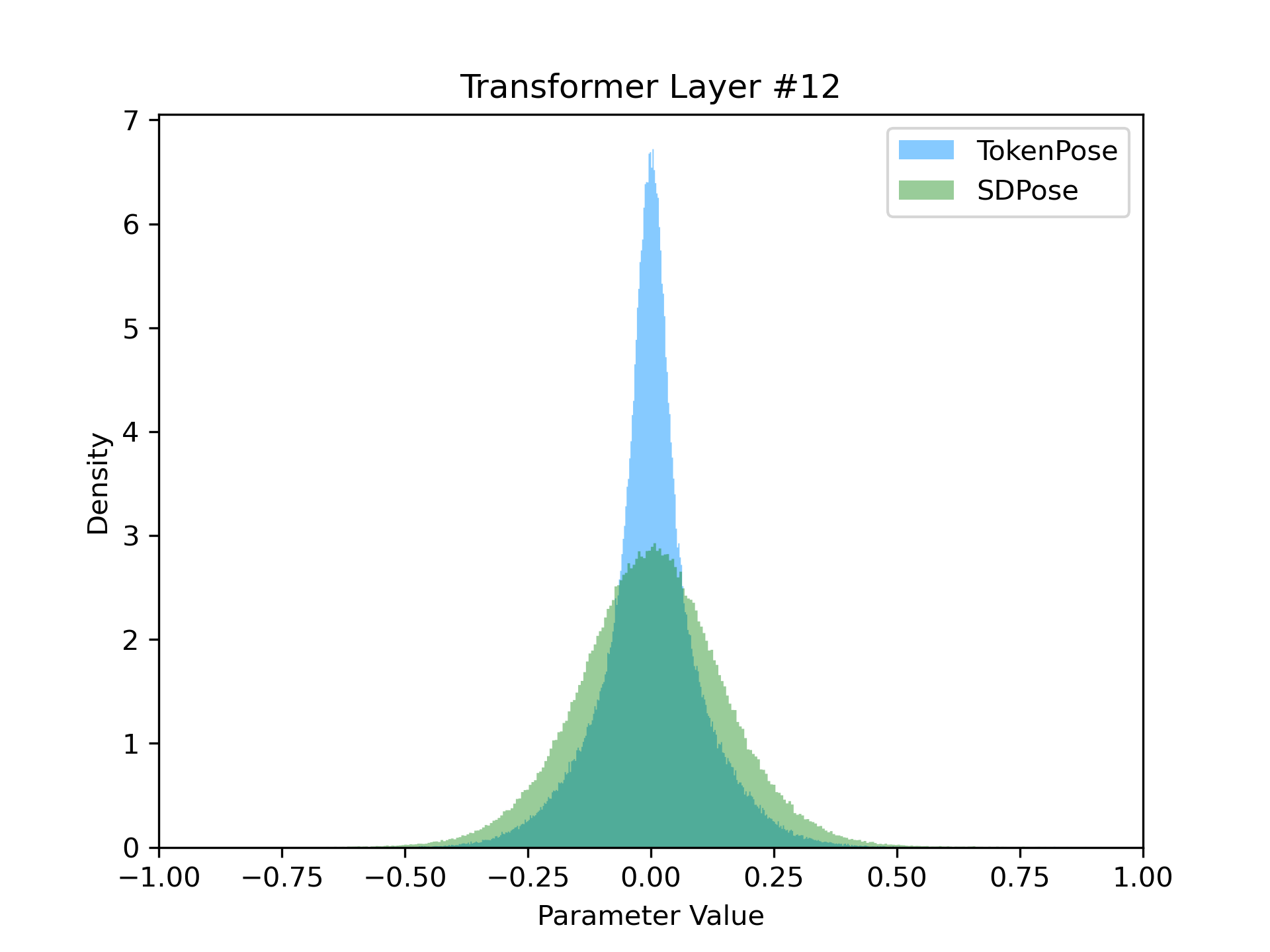}
    \caption[width=0.95\linewidth]{Visualization of parameter distributions for transformer layer \# 12. The blue represents TokenPose-S-V1\cite{Tokenpose} and the green represents SDPose-S-V1. There are fewer parameters close to 0 in our method, which proves that the parameters are more fully learned.}
    \label{fig:param_distri}
    \vspace{-0.6cm}
\end{figure}

    To help small models learn better, one possible way is to distill knowledge from big model to small model\cite{OKDHP,distilpose}. 
    %
    However, previous distillation methods have the following drawbacks: 
    (1) To align the vector space, an additional manipulation is required during feature distillation\cite{vitkd} and leads to a potential performance decrease.
    (2) A huge extra training cost is required to train the powerful teacher network.
     
    %
    %
    %
    %
    %
    %
    %
    %
    %
    %
    
    In this paper, we introduce a cyclic forwarding scheme, for which we further design a self-distillation method.
    This framework, termed SDPose, mitigated the conflicts between scale and performance for HPE. 
    The key insight guiding our designs is that for a deep HPE method, its performance can improve proportionally to what we define as the model's \textit{latent depth}.
    \textit{Latent depth} is the transformer layers depth involved in the complete inference process.
    Adding layers to the model is the straightforward way to increase \textit{latent depth}, 
    but it also incurs extra parameters.
    To increase the \textit{latent depth} without adding extra parameters,
    we design the Multi-Cycled Transformer(MCT) module, which passes the tokenized features through the transformer layers in multiple cycles during inference and uses the last output as the result.
    %
    %
    As shown in Fig. \ref{fig:param_distri}, compared to the transformer-based models,
    the parameters of the MCT-based models have higher variance and 
    lower density near zero, 
    which proves that it has been better trained.
    %
    %
    In this way, utilizing the MCT module can help small transformer-based models to be considered as a transformer-based model with greater \textit{latent depth}, 
    and break free from under-fitting to achieve a better performance. 

    Nevertheless, the MCT module still adds extra computational effort.
    In order to avoid additional computational consumption,
    we come up with a quite simple but effective self-distillation paradigm.
    Specifically, during the training phase, we send the tokenized features into the MCT module, 
    and because the input and output are in the same vector space for each cycle in the MCT module, 
    previous results can be distilled from the latter outputs without any additional operations.
    %
    At inference time, we perform one single pass to maintain the original computation consumption.
    With this design, we extract the knowledge of the MCT module into a naive forward model in one training, resulting in a better-trained model.
    Overall, our method achieved improved performance while maintaining the computational consumption.

    We designed several SDPose models based on TokenPose\cite{Tokenpose} and DistilPose\cite{distilpose}: SDPose-T, SDPose-S-V1, SDPose-S-V2, SDPose-B and SDPose-Reg.
    As evident in Fig. \ref{fig:parameter_ap}, our MCT-based models achieved improved performance under the same compute consumption as their base models. They also achieved similar performance compared to other much larger models.

    Our contribution can be summarized as follows:
    \begin{itemize}
        \item[$\bullet$]We are the first to find that looping the token through the transformer layers can increase the \textit{latent depth} of the transformer layers without adding extra parameters. 
        Based on this finding, we design Multi-Cycled Transformer(MCT) module.
        \item[$\bullet$]We design a self-distillation paradigm SDPose that extracts the knowledge in the MCT module into one single pass model, achieving a balance between performance and resource consumption. 
        To the best of our knowledge, we are the first to explore how self-distillation can be applied to the transformer-based HPE task.
        \item[$\bullet$]We have conducted extensive experiments and analyses to demonstrate the effectiveness and broad applicability of our approach on multiple tasks.
    \end{itemize}

\begin{figure*}[t]
    \vspace{-0.5cm}
    \centering
    \includegraphics[width=1\textwidth]{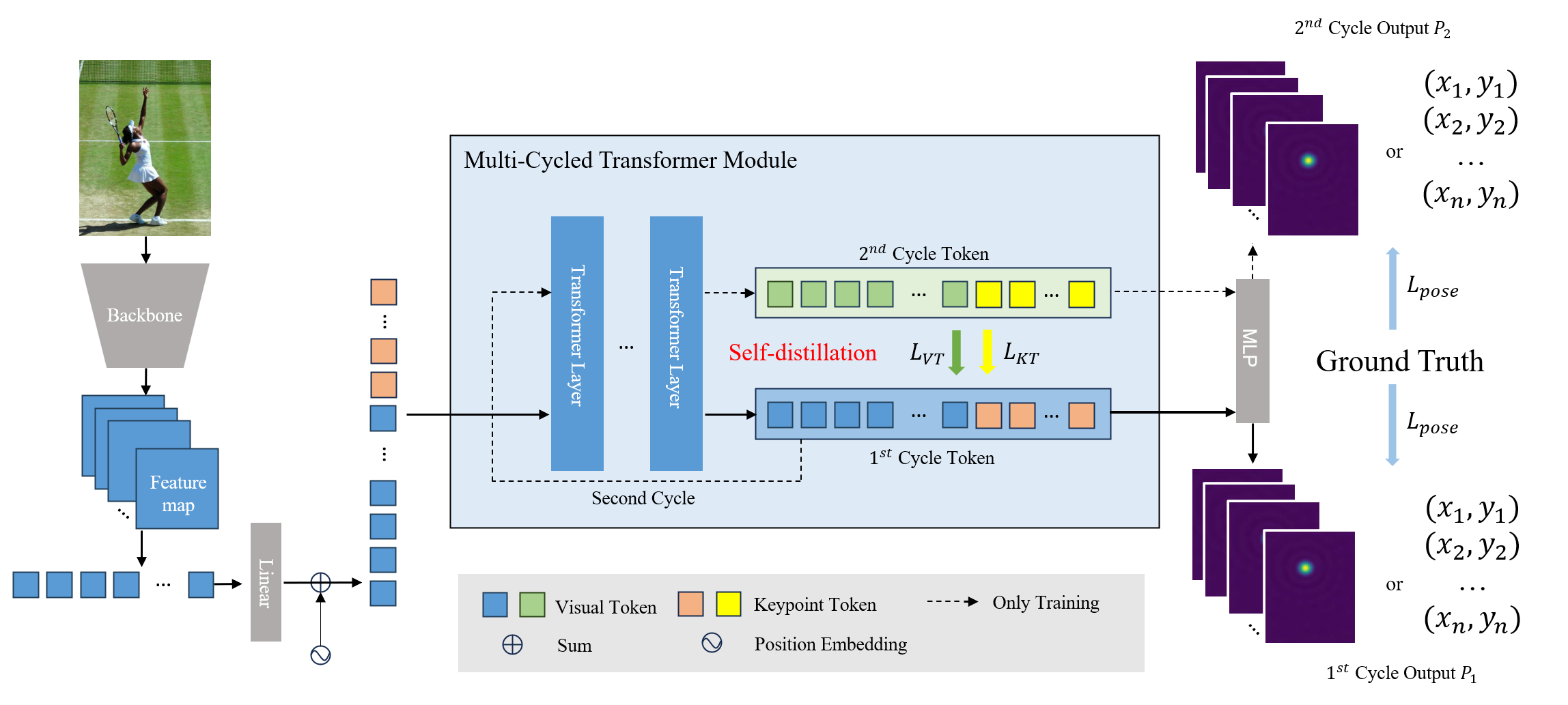}
    \caption[width=0.95\textwidth]{Overall architecture of SDPose used twice cycles. During training, The visual tokens and keypoint tokens will pass through the transformer layers twice. The tokens and heatmaps obtained during the second time will serve as the teacher to distill the tokens and heatmaps obtained during the first time.}
    \vspace{-0.4cm}
    \label{fig:pipeline}
\end{figure*}

\section{Related Works}

\subsection{Human Pose Estimation}

    Deep learning based methods dominate the HPE tasks.
    Deep learning based human pose estimators can be classified into regression-based and heatmap-based. 
    
    Regression-based methods directly estimate the coordinates for each keypoint. 
    Toshev \textit{et al.} \cite{deeppose} first leveraged a convolutional network to predict the image coordinates of 2D human joints, followed by numerous innovations in network architecture. Notably, transformer-based networks such as Poseur \cite{poseur} have achieved good prediction quality. 
    Besides seeking better network architectures, works like RLE \cite{rle} improved the regressor learning framework by quantifying the uncertainty of the regressed coordinates. 

    Heatmap-based methods estimate a 2D image or 3D volume of likelihood and decode it into coordinates. Since the seminal work by Tompson \textit{et al.}\cite{heatmap}, heatmap has become the predominant output representation, as its dimensionality better aligns with the input image space and thus reduces the learning complexity for neural networks.   

    For both the output representations, transformer has proven to be an effective module for feature extraction and processing for human pose estimators. 
    Yang \textit{et al.} \cite{transpose} utilizes transformer encoder to further encode the feature map produced by a convolutional neural network into keypoint representations. Xu \textit{et al.} \cite{vitpose}, on the other hand, designs a pure-transformer architecture for initial image feature extraction as well as feature processing. Li \textit{et al.} \cite{Tokenpose} designs token representation of keypoint information and feeds the learnable keypoint tokens as input to the transformer modules. While accurate, these transformer-based models tend to be complex. 
    
    Several works have proposed more lightweight transformer-based HPE models. Ye \textit{et al.} \cite{distilpose} designed DistilPose with a novel simulated heatmap loss to enable knowledge transfer from a heatmap-based teacher network to a regression-based student network. While it achieved SOTA performance, DistilPose requires training a teacher network for the separate distillation process. In comparison, Ma \textit{et al.} proposed PPT \cite{ppt} that finds and discards the less attended image token in TokenPose to reduce computational complexity. Although PPT reduced computation without extra hassles, it comes with the cost of performance drops compared to TokenPose. Previous works failed to reduce computation complexity while achieving SOTA results through an integral process, and our work is to our knowledge the first success towards this goal for HPE.

\subsection{Knowledge Distillation}
    To reduce the cost of training and deploying deep learning models, several techniques have been proposed, among which knowledge distillation is the most relevant to our method.

    Originally proposed by Hinton et.al \cite{distilling} as a model compression technique, knowledge distillation transfers knowledge from a teacher model to a student model. Recent works explored knowledge distillation within one model, namely self-distillation. Be Your Own Teacher \cite{beyourownteacher} distill knowledge in deeper layers into shallower layers within one model. Born-Again Neural Network \cite{bornagain} applies self-distillation along the temporal dimension, distilling knowledge from the model in previous iterations to supervise model learning in the current iteration. 
    Our work furthered this line of work by making the first effort to apply self-distillation on transformer-based HPE models.

\begin{figure*}[t]
    \vspace{-0.7cm}
    \centering
    \includegraphics[width=1\textwidth]{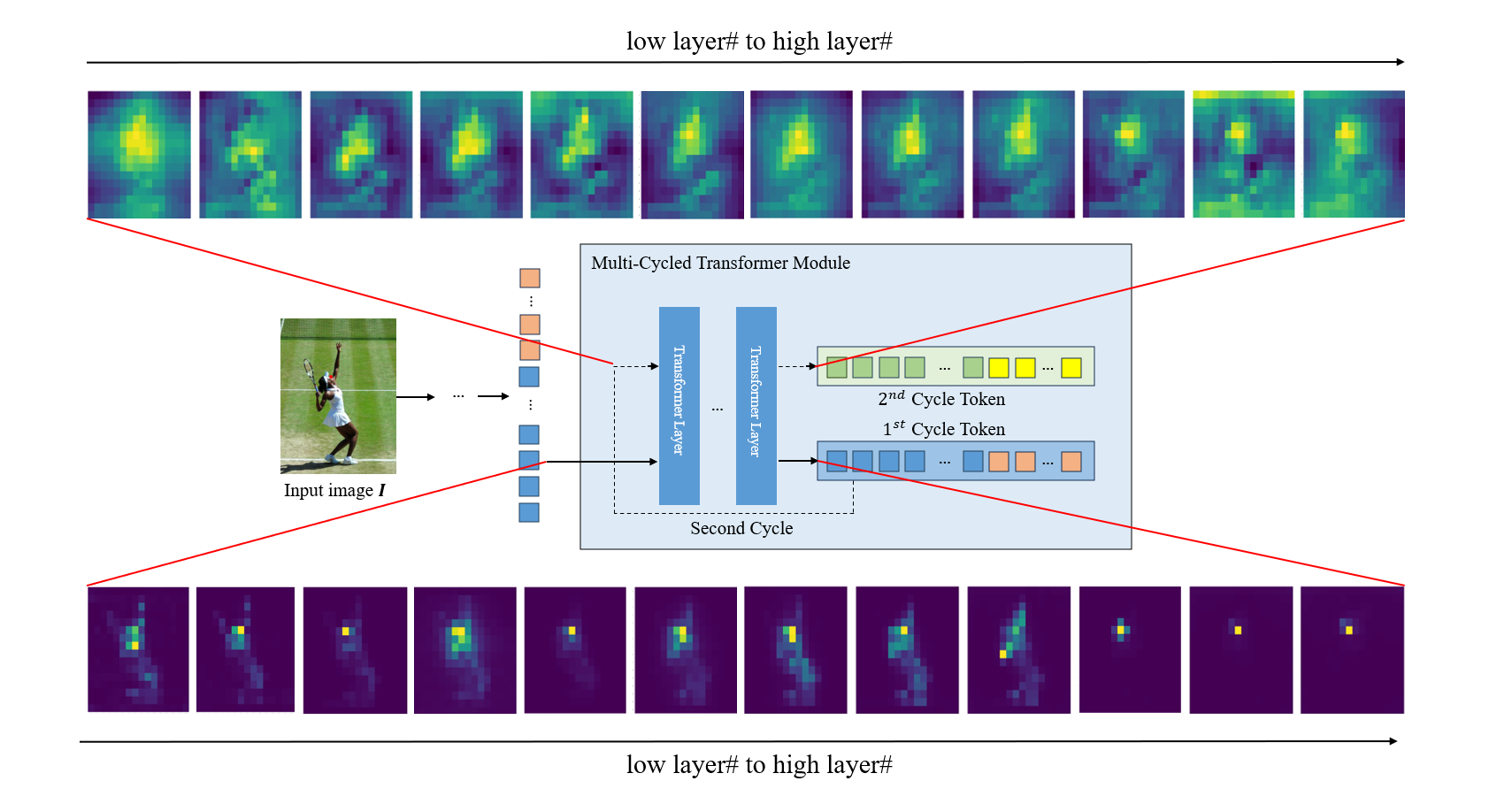}
    \caption[width=0.95\textwidth]{Visualization of the attention maps between nose keypoint token and visual tokens in different layers of MCT module. The lower one is from the first cycle, The top one is from the second cycle.}
    \label{fig:atten_vis}
    \vspace{-0.3cm}
\end{figure*}

\vspace{-0.25cm}
\section{Methods}
    In this section, we propose the Multi-Cycled transformer(MCT) module for our cyclic forwarding scheme.
    Further, we propose a self-distillation human pose estimation framework SDPose for our MCT module. 
    During training, the model passes tokens through the MCT module for several cycles,
    where the output from the previous cycle is used as the input to the next cycle.
    Then, we use the output of each cycle in the MCT module to distill the output of the previous cycle, 
    thus extracting the knowledge from the complete inference of the MCT module into one single pass.
    During inference, the model maintains its original inference pipeline, without incurring additional computation but achieving stronger performance. 
    The overall framework is shown in Fig. \ref{fig:pipeline}.

    \subsection{Multi-Cycled Transformer Module}
    To better improve small transformer-based models, we first investigated how to go about increasing the \textit{latent depth} of small transformer-based models.
    We propose the MCT module, 
    which loops tokenized features multiple cycles through the transformer layers and makes the performance of the transformer network equivalent to that of a deeper transformer network.
    
    Specifically, we followed the scheme of TokenPose\cite{Tokenpose}.
    for an input image \emph{I}, we extract feature \emph{F} from the backbone and then divide \emph{F} into a grid of patches.
    Then we flatten them and use a linear projection function to form them as visual tokens \emph{VT}. 
    And we use extra \emph{K} learnable tokens \emph{KT} to represent \emph{K} keypoints.
    Then we concatenated keypoint tokens with visual tokens and sent them to transformer encoder layers.
    For a MCT module that of cycled \emph{N} times, we denote the output keypoint tokens and visual tokens of each cycle as ${VT}_{1}, {KT}_{1}...{VT}_{N}, {KT}_{N}$, respectively.
    For the $i^{th}$ cycle, we take ${VT}_{i-1}$ and ${KT}_{i-1}$ as inputs and the outputs is ${VT}_{i}$ and ${KT}_{i}$.
    At last, we use ${VT}_{N}$ and ${KT}_{N}$ as the transformer layers' output to make the prediction.
    The MCT module we designed has higher \textit{latent depth} compared with transformer model which has the same number of transformer layers, 
    and allows the parameters to be learned more fully for better performance. 
    We further interpret this conclusion in Sec. \ref{sec:visualization}.

\begin{table*}[t]
\centering
\begin{tabular}{l|l|l|llllll}
\hline
\multirow{2}{*}{Method}            & \multirow{2}{*}{Params(M)} & \multirow{2}{*}{GFLOPs} & \multirow{2}{*}{\emph{AP}} & \multirow{2}{*}{\emph{$AP^{50}$}} & \multirow{2}{*}{\emph{$AP^{75}$}} & \multirow{2}{*}{\emph{$AP^{M}$}} & \multirow{2}{*}{\emph{$AP^{L}$}} & \multirow{2}{*}{\emph{AR}} \\ 
 & & & & & & & & \\ \hline
\specialrule{0em}{1pt}{1pt}
SimpleBaseline-Res50\cite{simplebaseline}       & 34.0               & 8.9             & 70.4        & 88.6          & 78.3          & 67.1         & 77.2         & 76.3        \\
SimpleBaseline-Res101\cite{simplebaseline}      & 53.0               & 12.4            & 71.4        & 89.3          & 79.3          & 68.1         & 78.1         & 77.1        \\
SimpleBaseline-Res152\cite{simplebaseline}      & 68.6               & 15.7            & 72.0        & 89.3          & 79.8          & 68.7         & 78.9         & 77.8        \\
TokenPose-S-V1*\cite{Tokenpose}            & 6.6$\dagger$       & 2.4$\dagger$    & 69.5$\dagger$         & 87.7          & 77.1          & 65.7         & 76.6         & 74.9        \\
TokenPose-S-V2*\cite{Tokenpose}            & 6.2                & 4.7            & 71.8$\ddagger$        & 88.7          & 79.0          & 68.3         & 78.5         & 77.0        \\
TokenPose-B*\cite{Tokenpose}               & 13.2               & 5.2             & 73.2$\S$        & 89.5          & 80.2          & 70.1         & 79.8         & 78.7        \\\hline
\specialrule{0em}{1pt}{1pt}
OKDHP-2HG\cite{OKDHP}                  & 13.0               & 25.5            & 72.8        & 91.5          & 79.5          & 69.9         & 77.1         & 75.6        \\
OKDHP-4HG\cite{OKDHP}                  & 24.0               & 47.0            & 74.8        & 92.5          & 81.6          & 72.1         & 78.5         & 77.4        \\
PPT-S*\cite{ppt}                     & 6.6                & 2.0             & 69.8        & 87.7          & 76.8          & 66.1         & 76.7         & 75.1        \\
PPT-B*\cite{ppt}                     & 13.2               & 4.7             & 73.4        & 89.5          & 80.8          & 70.3         & 79.8         & 78.8        \\\hline
\specialrule{0em}{1pt}{1pt}
\textbf{SDPose-T(Ours)}          & 4.4$\dagger(\textcolor{red}{\downarrow 33.3\%})$   & 1.8$\dagger(\textcolor{red}{\downarrow 25.0\%})$    & \textbf{69.7}$\dagger(\textcolor{red}{\uparrow 0.2\%})$   & \textbf{88.1}         & \textbf{77.3}          & \textbf{66.1}         & \textbf{76.6}         & \textbf{75.2}        \\
\textbf{SDPose-S-V1(Ours)}       & 6.6                & 2.4             & \textbf{72.3}$\dagger(\textcolor{red}{\uparrow 2.8\%})$        & \textbf{89.2}          & \textbf{79.6}          & \textbf{68.8}         & \textbf{79.1}         & \textbf{77.7}        \\
\textbf{SDPose-S-V2(Ours)}       & 6.2                & 4.7            & \textbf{73.5}$\ddagger(\textcolor{red}{\uparrow 1.7\%})$       & \textbf{89.5}          & \textbf{80.4}          & \textbf{70.1}         & \textbf{80.3}         & \textbf{78.7}        \\
\textbf{SDPose-B(Ours)}          & 13.2               & 5.2            & \textbf{73.7}$\S(\textcolor{red}{\uparrow 0.5\%})$        & \textbf{89.6}          & \textbf{80.4}          & \textbf{70.3}         & \textbf{80.5}         & \textbf{79.1}        \\ \hline
\end{tabular}
\caption{Results of heatmap-based methods on MSCOCO validation dataset. the input size is 256$\times$192. * means we re-train and evaluate the models on mmpose\cite{mmpose2020}. $\dagger,\ddagger$ and $\S$ represents the data pair for comparison.}
\vspace{-0.1cm}
\label{tab:table1}
\end{table*}

\begin{table*}[t]
\centering
\begin{tabular}{l|l|l|l|l|l|l}
\hline
\specialrule{0em}{1pt}{1pt}
Methods                   & Backbone      &Input Size       & Params(M) & GFLOPs & FPS  & \textit{AP} \\ \hline
\specialrule{0em}{1pt}{1pt}
PRTR\cite{cascade}                      & ResNet-50     &384$\times$288   & 41.5      & 11.0 & 33.8  & 68.2        \\
PRTR\cite{cascade}                     & ResNet-50     &512$\times$384   & 41.5      & 18.8 &  32.7      & 71.0        \\
Poseur\cite{poseur}                    & MobileNetV2   &256$\times$192   & 11.4      & 0.5   & 12.1   & 71.9        \\
Poseur\cite{poseur}                    & ResNet-50     &256$\times$192   & 33.3      & 4.6   & 12.0   & 75.4      \\
RLE\cite{rle}                       & ResNet-50     &256$\times$192   & 23.6      & 4.0      & 51.5       & 70.5        \\
DistilPose-S\cite{distilpose}              & Stemnet       &256$\times$192   & 5.4   & 2.4   & 39.2   & 71.6        \\ \hline
\specialrule{0em}{1pt}{1pt}
\textbf{SDPose-Reg(Ours)} & Stement       &256$\times$192   & 5.4       & 2.4  & 38.3  & \textbf{73.0}       \\ \hline
\end{tabular}
\caption{Results of regression-based methods on MSCOCO validation dataset.}
\vspace{-0.4cm}
\label{tab:table2}
\end{table*}

    \subsection{Self-Distillation On MCT Module}

    The MCT module incurs additional computation, which we want to avoid without sacrificing model performance. 
    
    A seemingly promising approach is to use the result of the first cycle in the MCT directly, but this will drastically reduce performance.
    As shown in Fig.~\ref{fig:atten_vis}, during the first cycle through the transformer layers, the attention of the keypoint tokens is consistently focused on a smaller area and gradually contracts to a single location. 
    However, during the second cycle through the transformer layers, the attention of the keypoint tokens expands to a larger area across all layers. 
    This demonstrates that each cycle carries rich information, 
    and naively ignoring outputs from latter cycles lead to information loss and thus performance degradation.
    
    Inspired by works in self-distillation, we use the complete inference in the MCT module as a teacher and extract the knowledge of it into one single pass in the MCT module, which we used as a student.
    Since the input and output tokens are in the same vector space in the transformer layers, we can distill between output tokens from different cycles with minimized information loss. 

    Specifically, for an MCT module cycled \emph{N} times, 
    we use the output ${VT}_{i}$ and ${KT}_{i}$ to distill the the output ${VT}_{i-1}$ and ${KT}_{i-1}$ in the previous cycle.
    During training, with each inference we distill all the cycles, thus gradually distilling the knowledge to the first cycle.
    
    Meanwhile, in order to constrain the correctness of the tokens, 
    we send all the output tokens ${VT}_{1},{KT}_{1}...{VT}_{N},{KT}_{N}$ through the same prediction head to get predicted result ${P}_{1},{P}_{2}...{P}_{N}$ respectively and constrain the predictions with the ground-truth.

    \subsection{Loss Function}
    
    For the tokens of each cycle, we use the output tokens as teachers to distill the input tokens, respectively. Specifically, our loss is designed as:


    \begin{equation}
        L_{kt} = \sum_{i=1}^{N-1} MSE\left({KT}_{i},{KT}_{i+1}\right)
    \label{eq:keypoint_distil_loss} 
    \end{equation}


    \begin{equation}
        L_{vt} = \sum_{i=1}^{N-1} MSE\left({VT}_{i},{VT}_{i+1}\right)
    \label{eq:vision_distil_loss} 
    \end{equation}
    
    \noindent where $MSE$ refers to the Mean Squared Error Loss which has been shown to be effective in measuring differences between tokens. 

    Meanwhile, in order to ensure that the results of both cycle predictions are correct, we compute the loss of both predicted heatmaps with the ground truth:


    \begin{equation}
        L_{pose} = \sum_{i=1}^{N} MSE\left({P}_{i},GT \right)
    \label{eq:gt_loss} 
    \end{equation}

    \noindent where GT represents the ground truth.

    In summary, The overall loss function of our self-distillation framework is as follows:

    \begin{equation}
    L = L_{pose} + \alpha_1 L_{kt} + \alpha_2 L_{vt},
    \label{eq:sum_loss}
    \end{equation}

    \noindent where $\alpha_1, \alpha_2$ are hyper-parameters.

\begin{table*}[t]
\centering

\begin{tabular}{lllllllll}
\hline
\specialrule{0em}{1pt}{1pt}
\multicolumn{1}{l|}{Methods}                    & \multicolumn{1}{l|}{Input Size}       & \multicolumn{1}{l|}{Params(M)} & \multicolumn{1}{l|}{GFLOPs} & \textit{AP} & \textit{$AP^{50}$} & \textit{$AP^{75}$} & \textit{$AP^{M}$} & \textit{$AP^{L}$} \\ \hline
\specialrule{0em}{1pt}{1pt}
\multicolumn{9}{l}{\textit{\textbf{heatmap based methods}}}                                                                                                                                                                       \\ \hline
\specialrule{0em}{1pt}{1pt}
\multicolumn{1}{l|}{TokenPose-S-V1*\cite{Tokenpose}}            & \multicolumn{1}{l|}{256 $\times$ 192} & \multicolumn{1}{l|}{6.6}       & \multicolumn{1}{l|}{2.4}    & 68.6        & 89.9          & 76.1          & 65.1         & 74.5         \\
\multicolumn{1}{l|}{TokenPose-S-V2*\cite{Tokenpose}}            & \multicolumn{1}{l|}{256 $\times$ 192} & \multicolumn{1}{l|}{6.6}       & \multicolumn{1}{l|}{4.7}    & 71.1        & 90.4          & 78.7          & 67.7         & 77.1         \\
\multicolumn{1}{l|}{PPT-S*\cite{ppt}}                     & \multicolumn{1}{l|}{256 $\times$ 192} & \multicolumn{1}{l|}{6.6}       & \multicolumn{1}{l|}{2.0}    & 69.2      & 90.1        & 76.8        &  65.8      &      75.2  \\
\multicolumn{1}{l|}{\textbf{SDPose-T(Ours)}}    & \multicolumn{1}{l|}{256 $\times$ 192} & \multicolumn{1}{l|}{4.4}       & \multicolumn{1}{l|}{1.8}    & 69.2        & 90.2          & 76.8          & 65.7         & 75.2         \\
\multicolumn{1}{l|}{\textbf{SDPose-S-V1(Ours)}} & \multicolumn{1}{l|}{256 $\times$ 192} & \multicolumn{1}{l|}{6.6}       & \multicolumn{1}{l|}{2.4}    & 71.7        & 91.1          & 79.5          & 68.3         & 77.5         \\
\multicolumn{1}{l|}{\textbf{SDPose-S-V2(Ours)}} & \multicolumn{1}{l|}{256 $\times$ 192} & \multicolumn{1}{l|}{6.2}       & \multicolumn{1}{l|}{4.7}    & \textbf{72.7}        & \textbf{91.2}          & \textbf{80.3}          & \textbf{69.3}         & \textbf{78.5}         \\ \hline
\specialrule{0em}{1pt}{1pt}
\multicolumn{9}{l}{\textit{\textbf{regression based methods}}}                                                                                                                                                                     \\ \hline
\specialrule{0em}{1pt}{1pt}
\multicolumn{1}{l|}{PRTR-Res101\cite{cascade}}                & \multicolumn{1}{l|}{512 $\times$ 384} & \multicolumn{1}{l|}{60.4}      & \multicolumn{1}{l|}{33.4}   & 72.0        & 89.3          & 79.4          & 67.3         & 79.7         \\
\multicolumn{1}{l|}{RLE-Res50*\cite{rle}}                 & \multicolumn{1}{l|}{256 $\times$ 192} & \multicolumn{1}{l|}{23.6}      & \multicolumn{1}{l|}{4.0}    & 69.8        & 90.1          & 77.5          & 67.2         & 74.3         \\
\multicolumn{1}{l|}{DistilPose-S*\cite{distilpose}}                 & \multicolumn{1}{l|}{256 $\times$ 192} & \multicolumn{1}{l|}{5.4}      & \multicolumn{1}{l|}{2.4}    & 71.0        & 91.0          & 78.9          & 67.5         & 76.8         \\
\multicolumn{1}{l|}{\textbf{SDPose-Reg(Ours)}}  & \multicolumn{1}{l|}{256 $\times$ 192} & \multicolumn{1}{l|}{5.4}       & \multicolumn{1}{l|}{2.4}    &  \textbf{72.1}           &    \textbf{91.2}           &   \textbf{79.5}            &   \textbf{68.6}           &    \textbf{78.0}          \\ \hline
\end{tabular}
\caption{Result on MSCOCO test-dev dataset. * means we re-train and evaluate the models on MMPose\cite{mmpose2020}.}
\vspace{-1.5em}
\label{tab:table3}
\end{table*}

\begin{table}[t]
\centering
\scalebox{0.85}{
\begin{tabular}{l|c|c|c|c}
\hline
\specialrule{0em}{1pt}{1pt}
Methods         & AP      & AR   &Params(M) &GFLOPs    \\ \hline
\specialrule{0em}{1pt}{1pt}
SimpleBaseline-Res50  & 63.7  & 73.2 & 34.0 & 8.9\\
KAPAO-S\cite{kapao}  & 63.8  & - & 12.6 & - \\
PPT-S\cite{ppt}  & 55.6  & 65.2 & 6.6 & 2.0 \\
RLE-Res50\cite{rle}  & 57.0  & 66.9 & 23.6 & 4.0 \\
\hline
\specialrule{0em}{1pt}{1pt}
\textbf{SDPose-S-V1}  & \textbf{57.3} & \textbf{66.8} & \textbf{6.6} & \textbf{2.4} \\
\textbf{SDPose-S-V2}  & \textbf{64.5} & \textbf{73.7} & \textbf{6.2} & \textbf{4.7} \\
\hline
\end{tabular}}
\caption{Results of heatmap-based methods on Crowdpose test dataset. $\dagger$ and $\ddagger$ represents the data pair for comparison.}
\vspace{-1.5em}
\label{tab:crowdpose}
\end{table}

\section{Experiments}

    We evaluated the proposed SDPose models and performed abundant ablation studies on MSCOCO as well as Crowdpose dataset\cite{mscoco,crowdpose}. 
    For fairness, all our experiments are conducted using MMPose\cite{mmpose2020} framework.

    \subsection{Implementation Details}
        \subsubsection{Datasets}
        We conducted experiments on 2 datasets, the MSCOCO dataset\cite{mscoco} and the Crowdpose dataset~\cite{crowdpose}. 
        MSCOCO contains over 200K human body images, with each human body having 17 pre-annotated keypoints. 
        We use MSCOCO train2017 with 57K images to train our models and compare methods. 
        We evaluated them on both MSCOCO val2017 with 5K images and dev2017 with 20K images, individually. 
        We also evaluated our methods in the more challenging crowded scene using Crowdpose. This dataset consists of 20K human body images containing about 80K persons with overlaps of body parts, with 14 pre-annotated keypoints per person.
        
        The bounding box and evaluation metrics used for our evaluations are consistent with previous works\cite{rle,Tokenpose, crowdpose}.

        \subsubsection{Settings And Training}
        We applied our method to TokenPose-S-V1, TokenPose-S-V2, and TokenPose-B. We substitute the naive transformer module with our MCT module and all other model configurations remain consistent with those in TokenPose\cite{Tokenpose} paper. We name the MCT-based models SDPose-S-V1, SDPose-S-V2, and SDPose-B.
        To demonstrate that our method can improve the \textit{latent depth}, we set up a smaller model SDPose-T which changes TokenPose-S-V1 to six transformer layers and inference three cycles during training, using the latter cycle to distill the former.
        Meanwhile, we also designed a regression-based model SDPose-Reg, which uses the regression-based head named TokenReg in Distilpose\cite{distilpose} with an RLE loss\cite{rle}, and used our method to train.
        For our method, we train the models on a machine with 8 NVIDIA Tesla V100 GPUs, allocating 64 samples per GPU. We use the Adam optimizer for 300 epochs of training. 
        The initial learning rate was set to 1e-3 and decayed by a factor of ten at epochs 200 and 260, respectively. 
        For our loss function, we set the hyper-parameters to $ \alpha_1 = \alpha_2 = 5e-6.$ 
        %

    \subsection{Main Results}
    
    \textbf{Compared with heatmap-based methods.} 
    As Tab. \ref{tab:table1} shown, our proposed SDPose achieves competitive performance compared with the other small-scale models.
    We mainly compare our methods with TokenPose\cite{Tokenpose}, OKDHP\cite{OKDHP}, and PPT\cite{ppt}. 
    Specifically, SDPose-S-V1 achieves 72.3$\%$ AP with 6.6M params and 2.4 GFLOPs, Which under the same parameter and computational complexity as TokenPose-S-V1, and makes a 2.8$\%$ AP improvement. 
    Similarly, SDPose-S-V2 and SDPose-B achieve 1.7$\%$ and 0.5$\%$ AP improvement with the same parameter and GFLOPs as TokenPose-S-V2 and TokenPose-B, respectively.
    Furthermore, SDPose-T slightly improves performance($\uparrow0.2\%$) with a significantly lower number of parameters($\downarrow33.3\%$) and GFLOPs($\downarrow25.0\%$) compared to TokenPose-S-V1.
    Compared to other lightweight methods, our approach achieved higher performance with fewer parameters in most cases. 
    Specifically, SDPose-T reduces more number of parameters and computation without degrading the performance compared to PPT-S. 
    Also, Tab. \ref{tab:table3} shows the results of our method and those of the other small models on the MSCOCO test-dev set.
    We see that SDPose-S-V2 achieved SOTA performance among the small models.
    In addition, our method can be applied to PPT to get better performance.
    Detail results are presented in Sec. \ref{sec:extensibily}.

    \noindent
    \textbf{Compared with regression-based methods.} 
    As shown in Tab. \ref{tab:table2} and Tab. \ref{tab:table3}, our proposed SDPose achieves competitive performance compared with the other regression models.
    Compared with PRTR\cite{cascade}, which is also a transformer-based model, our method achieved a 1.5$\%$ AP improvement with a significant reduction in the number of parameters and GFLOPs.
    Compared with the smaller Poseur\cite{poseur}, our method has a 0.6$\%$ AP improvement with a significant FPS improvement($\uparrow26.2$).
    Compared with DistilPose-S\cite{distilpose}, $1.4\%$ AP improvement was also obtained using our method.
    Also, as Tab. \ref{tab:table3} shown, SDPose-Reg makes a 1.1$\%$ AP improvement compared with DistilPose-S\cite{distilpose} on the MSCOCO test-dev set.

    \noindent
    \textbf{Evaluation on Crowdpose dataset.} 
    To verify our model's generalizability and to challenge our models to a harder scenario, we trained and evaluated our models on the Crowdpose dataset.  
    As shown in Tab. \ref{tab:crowdpose}, all of our MCT-based models outperform their corresponding naive transformer-based baselines.

\begin{table}[t]
\centering
\begin{tabular}{c|cc|cc}
\hline
\specialrule{0em}{1pt}{1pt}
\multirow{2}{*}{$L_{pose}$} & \multicolumn{2}{|c|}{Distillation} & \multirow{2}{*}{AP} & \multirow{2}{*}{Improv.} \\
    \cline{2-3}
     & $L_{kt}$     & $L_{vt}$     &      &           \\ \hline
\specialrule{0em}{1pt}{1pt}
            &             &             & 55.4\% & -                    \\
\hline
$\checkmark$ &              &              & 70.2\%  & $\uparrow14.8\%$                     \\
$\checkmark$ & $\checkmark$ &              & 69.6\%   & $\uparrow14.2\%$     \\
$\checkmark$ &              & $\checkmark$ & 71.7\% & $\uparrow16.3\%$     \\
             & $\checkmark$ & $\checkmark$ & 58.2\% & $\uparrow2.8\%$      \\ \hline
\specialrule{0em}{1pt}{1pt}
$\checkmark$ & $\checkmark$ & $\checkmark$ & 72.3\% & $\uparrow16.9\%$     \\ \hline
\end{tabular}
\caption{ Ablation studies for different distillation types. All ablation experiments are based on SDPose-S-V1, The combination of all distillation loss brings the best performance, which is our method. $L_{pose}$ Means use the results predicted by each cycle to calculate the loss or use only the last cycle. Improv. = Improvement.}
\label{tab:ablation-study-1}
\end{table}

    \subsection{Visualization}
    \label{sec:visualization}
\begin{figure}[t]
    \centering
    \includegraphics[width=1\linewidth]{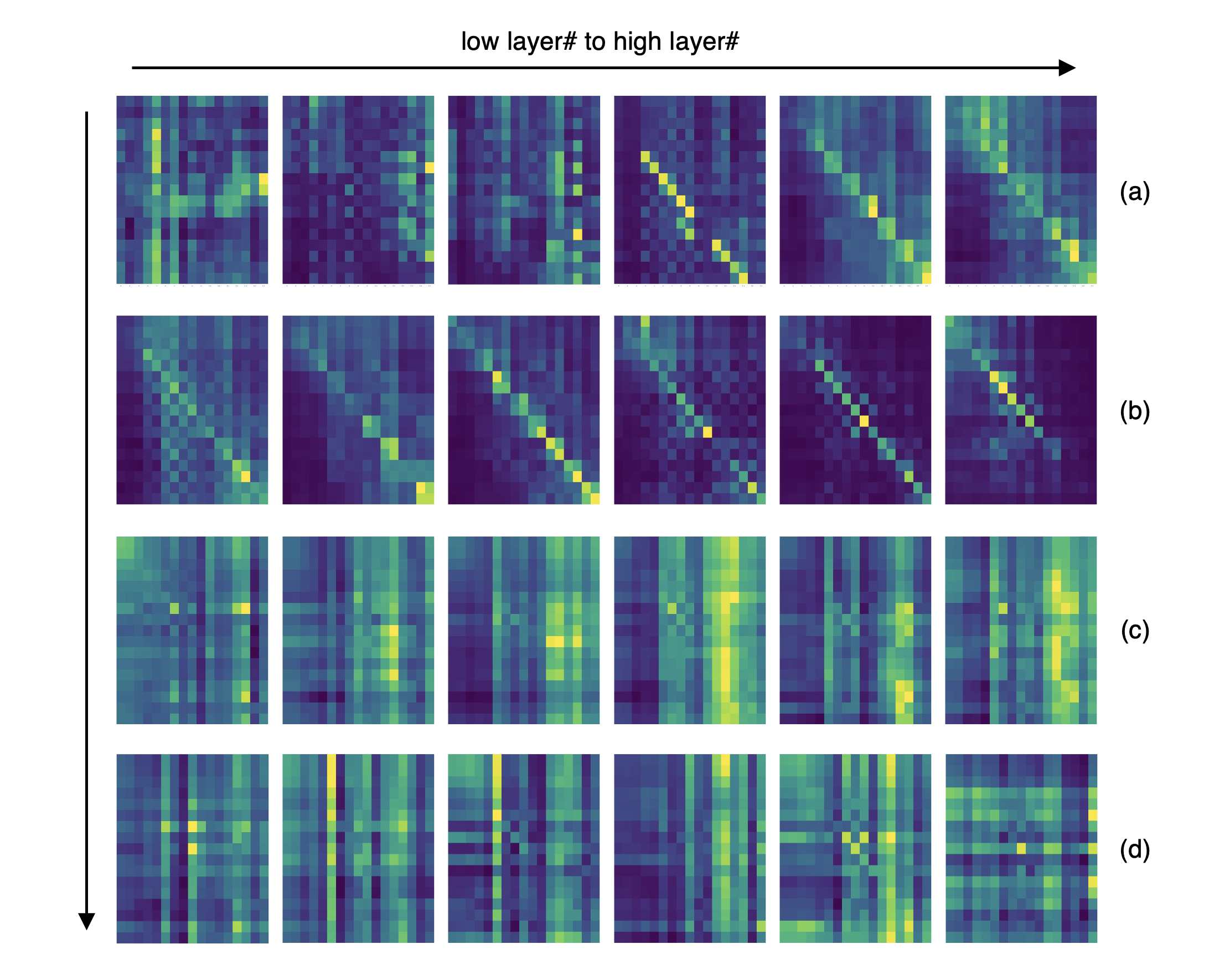}
    \caption[width=0.95\linewidth]{Visualization of the attention maps between nose keypoint token and visual tokens in different layers of MCT module. (a) (c) refer to layers \#1-\#6, (b) (d) refer to layers \#7-\#12 and (a) (b) refer to the first time, (c) (d) refer to the second time. }
    \label{fig:key_atten_vis}
\end{figure}

    To explore the reasons for the performance improvement of our method, we first visualized the attention map between keypoint tokens and visual tokens in different transformer layers of the MCT module. 
    As shown in Fig. \ref{fig:atten_vis}, each cycle carries a lot of information with it.
    We also visualized the attention maps between keypoint tokens in different transformer layers.
    As shown in Fig. \ref{fig:key_atten_vis}, during the first cycle through the transformer layers, 
    similar to the ordinary baseline method, the attention of the keypoint tokens gradually concentrates on themselves. 
    However, during the second cycle through the transformer layers, the attention of the keypoint tokens is redistributed to all other keypoint tokens, which we think represented obtaining more global information.
    Overall, this global information in the MCT module enables better learning of the parameters.

    Furthermore, we visualized the distribution of transformer parameters. 
    As shown in Fig. \ref{fig:param_distri}, TokenPose\cite{Tokenpose} has more near-zero parameters in each transformer layer, 
    which are commonly considered to be insufficiently trained parameters. 
    Our approach has significantly fewer near-zero parameters than TokenPose\cite{Tokenpose}, indicating that our network is more adequately trained. 
    This also demonstrates that the more global information contained in the MCT module allows for more adequate parameter learning.
    %
    %

\begin{table}[t]
\centering
\begin{tabular}{ccc|c}
\hline
\specialrule{0em}{1pt}{1pt}
Backbone  & Layers & Cycle   & AP                     \\ \hline
\specialrule{0em}{1pt}{1pt}
Stemnet  & 12     & 1       & 69.5                   \\
Stemnet  & 12     & 2       & 73.3\textcolor{red}{($\uparrow 3.8\%$)} \\
Stemnet  & 12     & 3       & 72.4\textcolor{red}{($\uparrow 2.9\%$)} \\
Stemnet  & 4      & 2       & 69.4\textcolor{red}{($\downarrow 0.1\%$)} \\
Stemnet  & 4      & 3       & 70.8\textcolor{red}{($\uparrow 1.3\%$)} \\\hline
\end{tabular}
\caption{Ablation studies for different cycle networks without distillation. Layers means the transformer layer number. Cycle means the number of times that token through transformer layers. }
\label{tab:ablation-study-2}
\end{table}

\begin{table}[t]
\centering
\begin{tabular}{cccc|c}
\hline
\specialrule{0em}{1pt}{1pt}
Backbone  & Layers & Cycle  & Distil.       & AP                     \\ \hline
\specialrule{0em}{1pt}{1pt}
Stemnet  & 12     & 1      & -             & 69.5                   \\
Stemnet  & 12     & 2      & 2→1           & 72.3\textcolor{red}{($\uparrow 2.8\%$)}  \\
Stemnet  & 12     & 3      & 3→2,2→1       & 71.7\textcolor{red}{($\uparrow 2.2\%$)}  \\ \hline
\end{tabular}
\caption{Ablation studies for different distillation settings. Layers means the transformer layer number. Cycle means the number of times that token through transformer layers. Distil. = distillation method, which means the type of distillation method. 2→1 means distillation second times result to first times result. }
\label{tab:ablation-study-3}
\end{table}

    \subsection{Ablation Studies}

        \subsubsection{Losses}
        In this section, we investigated the contribution of distillation losses from different parts to the performance of our method. 
        As shown in Tab. \ref{tab:ablation-study-1}, we set different distillation losses in various experiments. 
        When not using distillation and directly predicting results from the tokens obtained in the first cycle, the network only has the constraint from the second cycle's final output. 
        In this case, the tokens output from the first cycle are equivalent to intermediate results. 
        The direct use of its predictions loses the information embedded in the later cycles, so the performance is poor.
        When we use distillation loss, the model performance can be improved,
        which proves that the keypoint tokens learn more information through the MCT module and thus interact better with the model parameters.
        As we gradually add distillation losses, the performance of the first cycle prediction gradually improves. 
        When all three parts of the distillation loss are added, the performance reaches its best. 

        \subsubsection{Network Scale}
        In this section, We evaluated the effectiveness of our method under different network configurations and sizes. 
        As shown in Tab. \ref{tab:ablation-study-2}, We first investigated the effectiveness of our network augment method. 
        We set different MCT module cycle numbers based on TokenPose-S-V1 and trained without using self-distillation, predicting the output from the last cycle. 
        When we increase the number of cycles, the performance is improved compared to using only a single pass. 
        This fully demonstrates that our augment method enables the transformer layers to learn more information, and effectively augment the original network to a deeper transformer network. 
        To demonstrate the importance of introducing global information, we also designed a network with fewer layers but using three cycles.
        It is equivalent to going through 12 layers of transformer layers but gets better performance than the baseline.
        Furthermore, we noticed that the performance of the 12 transformer layers network with three cycles is lower than that of the network with two cycles. 
        We believe that although multiple cycles can help tokens pay attention to more global information, too many cycles may cause the network to forget the more critical local information of keypoints. 
        As shown in Tab. \ref{tab:ablation-study-3}, We designed a distillation experiment with a three-cycled MCT module.
        The test performance was lower than that of the distillation experiment with a two-cycled MCT module. 
        This also proves that excessive augment of the network structure to learn global knowledge is not entirely beneficial.

        %
        %
        %

    \subsection{Extensibility Study}

        In this section, we further investigate the extensibility of our approach. 
        As shown in Tab. \ref{tab:conbine-ppt}, we show the performance of our method when combined with PPT\cite{ppt}.
        For training, we pruned the tokens using PPT\cite{ppt} in the first cycle of SDPose.
        Ultimately, the performance maintained the level of SDPose-S-V1 while reducing the computation to the same of PPT-S.
        This proves that our method works well in conjunction with other lightweight methods.

        We also migrated SDPose to the classification task.
        We used Deit-Tiny\cite{deit} as our baseline.
        As shown in Tab. \ref{tab:classification}, when we applied the MCT module on Deit-Tiny\cite{deit}, the performance of the network was significantly improved. 
        When we trained the baseline using our SDPose, the performance of the network was also slightly improved without additional parameters and computations.
        This demonstrates the ability of our method to be extended to various tasks of the transformer-based model.
        
    \label{sec:extensibily}


\begin{table}[t]
\centering
\resizebox{\linewidth}{!}{
\begin{tabular}{c|cc|c}
\hline
\specialrule{0em}{1pt}{1pt}
Methods                                & Params(M) & GFLOPs   & AP                 \\ \hline
\specialrule{0em}{1pt}{1pt}
TokenPose-S-V1*\cite{Tokenpose}         & 6.6      & 2.4      & 69.5                   \\
PPT-S*\cite{ppt}                        & 6.6      & 2.0      & 69.8  \\
SDPose-S-V1+PPT\cite{ppt}               & 6.6      & 2.0      & 72.3  \\\hline
\end{tabular}}
\caption{Result of combined method on MSCOCO validation dataset. *means we re-train and evaluate the models on mmpose\cite{mmpose2020}. }
\label{tab:conbine-ppt}
\end{table}


\begin{table}[t]
\centering
\begin{tabular}{l|l}
\hline
\specialrule{0em}{1pt}{1pt}
Deit-Tiny\cite{deit}       & \multicolumn{1}{l}{AP}            \\ \hline
\specialrule{0em}{1pt}{1pt}
                & \multicolumn{1}{l}{72.2}          \\
+ 2 Cycle        & 73.4                              \\
+ SDPose (Ours) & \multicolumn{1}{c}{72.7\textcolor{red}{($\uparrow 0.5\%$)} } \\ \hline
\end{tabular}
\caption{Results on Imagenet 1K dataset. Cycle means we apply different cycle networks without
distillation on the Deit-Tiny.}
\label{tab:classification}
\end{table}

\section{Conclusion}

    In this work, we proposed a novel human pose estimation framework, termed SDPose, 
    which includes a Multi-Cycled Transformer(MCT) module and a self-distillation paradigm.
    Through our design, we have enabled the small transformer-based model to be dramatically improved without increasing the amount of computation and the number of parameters, 
    and achieved new state-of-the-art in the same scale models.
    Meanwhile, we also extend our method to other models, proving the generality of our method.
    
    In short, SDPose achieved state-of-the-art performance among the same scale models.


\section*{Acknowledgments}
    This work was supported by the National Natural Science Foundation of China (No.62302297, No.72192821), 
    Shanghai Municipal Science and Technology Major Project (2021SHZDZX0102), Shanghai Science and Technology Commission (21511101200), Shanghai Sailing Program (22YF1420300), Young Elite Scientists Sponsorship Program by CAST (2022QNRC001) and YuCaiKe[2023] Project Number: 14105167-2023.

\newpage

\bibliographystyle{unsrt}
\bibliography{main}

\end{document}


\maketitle

\section{Comparison with TokenPose-T}
We further investigated the performance of the proposed method when based on the TokenPose-T structure. 
TokenPose-T is a pure transformer-based variant. 
We compared the SDPose setting with its alignment named SDPose-T-V2 and the results on the MSCOCO test-dev dataset are shown in Tab. \ref{tab:tokenposet}.
Our approach also achieves significant performance improvements compared to the baseline.
This indicates that our approach is also applicable to a pure transformer-based model.

\begin{table}[h]
\centering
\begin{tabular}{l|l|l}
\hline
\specialrule{0em}{1pt}{1pt}
Methods                   & AP      &AR       \\ \hline
\specialrule{0em}{1pt}{1pt}
TokenPose-T*    & 60.5  &67.3 \\
\textbf{SDPose-T-V2}   & \textbf{63.1}\textcolor{red}{($\uparrow 2.6\%$)} & \textbf{73.7} \\
\hline
\end{tabular}
\caption{Results on MSCOCO test-dev dataset. * means we re-train and evaluate the models on MMPose.}
\label{tab:tokenposet}
\end{table}

\section{More Results On CrowdPose}
We further tested the results on Crowdpose based on SDPose-S-V1 with the version of the MCT module that uses three cycles, as shown in Tab. \ref{tab:3cycle}. 
Compared to SDPose-S-V1, SDPose-S-V1-3cycle gets better performance.
This demonstrates the importance of the global information introduced by the multiple cycles of the MCT module in more difficult tasks like Crowdpose. 

\begin{table}[h]
\centering
\begin{tabular}{l|l|l}
\hline
\specialrule{0em}{1pt}{1pt}
Methods                   & AP      &AR       \\ \hline
\specialrule{0em}{1pt}{1pt}
TokenPose-S-V1    & 55.7  &65.2 \\
\textbf{SDPose-S-V1}   & \textbf{57.3} \textcolor{red}{($\uparrow 1.6\%$)} &\textbf{66.8} \\
\textbf{SDPose-S-V1-3cycle}   & \textbf{57.5} \textcolor{red}{($\uparrow 1.8\%$)} &\textbf{67.0} \\
\hline
\end{tabular}
\caption{Results on Crowdpose test dataset.}
\label{tab:3cycle}
\end{table}

\section{More Visualization Results}
\begin{figure}[t]
    \centering
    \includegraphics[width=1\linewidth]{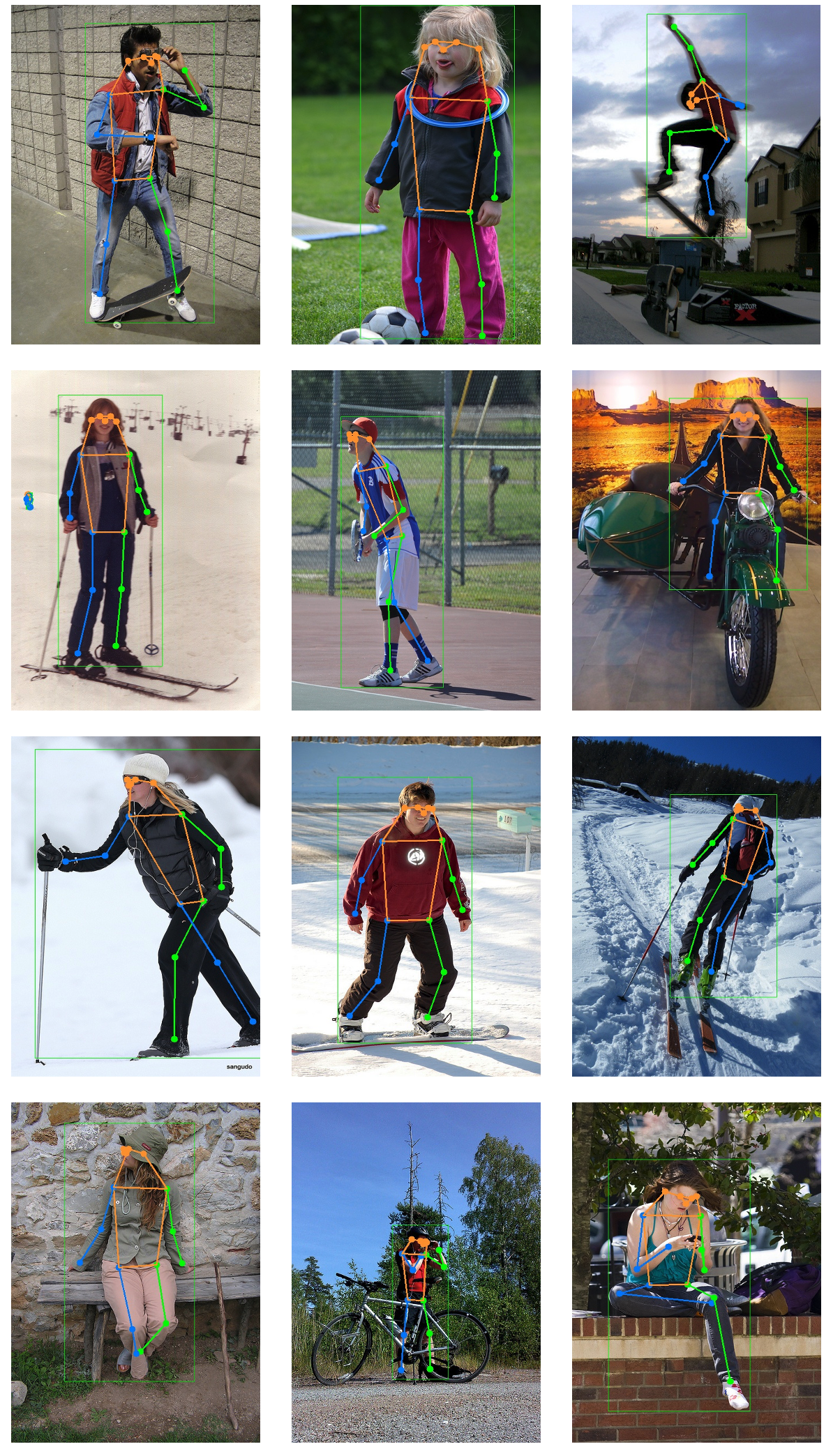}
    \caption[width=0.95\linewidth]{Visualization results of human pose estimation by SDPose-S-V1 on MSCOCO validation dataset.}
    \label{fig:visualization}
\end{figure}

We provide more visualization cases of Human Pose Estimation by SDPose-S-V1, as shown in Fig. \ref{fig:visualization}. 

\begin{figure*}[t]
    \centering
    \includegraphics[width=1\linewidth]{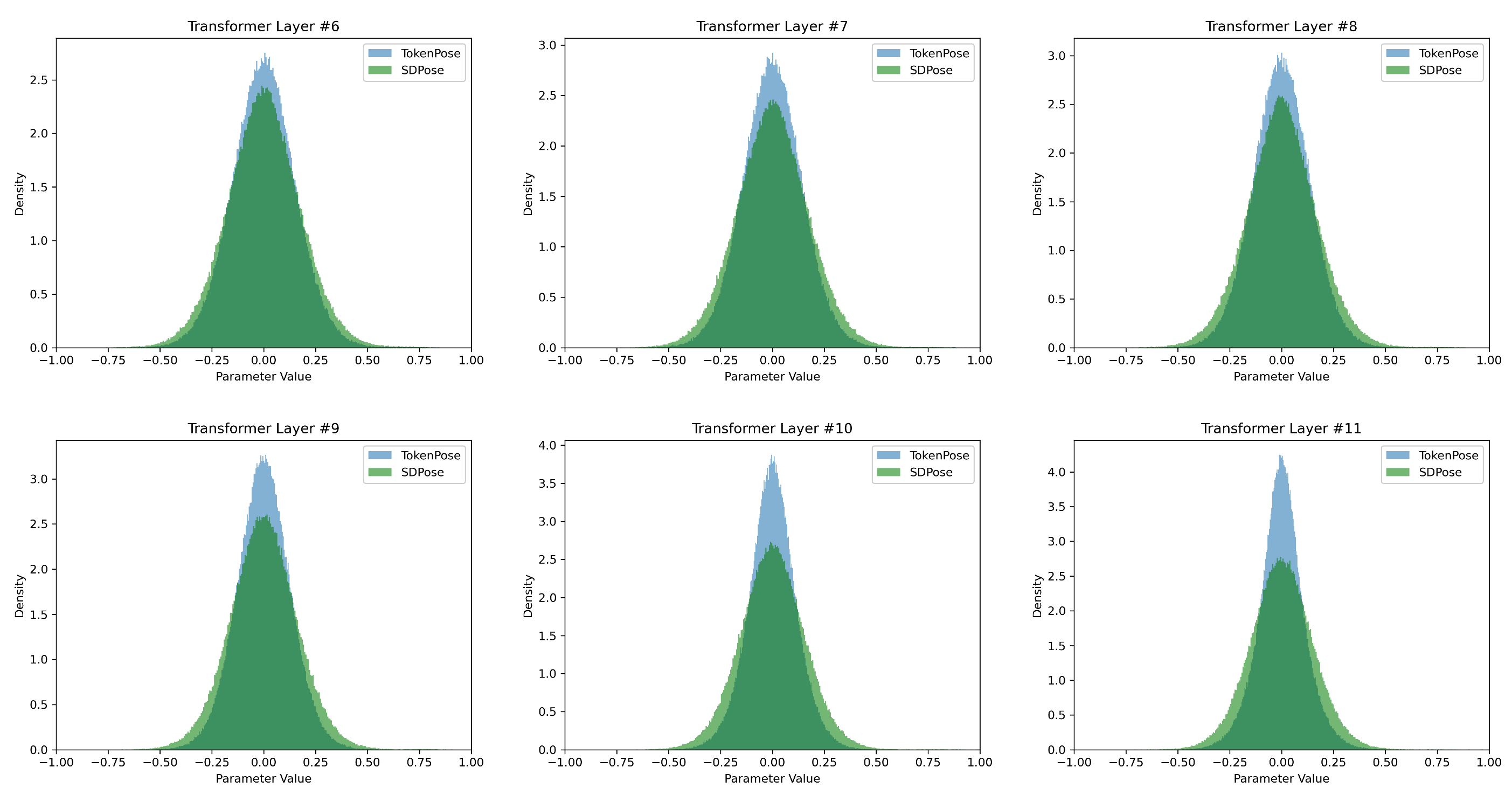}
    \caption[width=0.95\linewidth]{Visualization of parameter distributions for transformer layers \# 6 - \# 11. The blue represents TokenPose-S-V1 and the green represents SDPose-S-V1.}
    \label{fig:all_parameter}
\end{figure*}

\section{More Parameter Distributed Visualization Results}
We provide more visualization cases of parameters distributed in SDPose-S-V1, as shown in Fig. \ref{fig:all_parameter}. 
All parameter distributions of SDPose-S-V1 in Fig. \ref{fig:all_parameter} have a larger variance, proving that our method can fully optimize the whole transformer. 
Meanwhile, we note that layers at deeper locations are better optimized. 
This may represent richer global information learned by deeper layers.

{\small
	\bibliographystyle{ieee_fullname}
	\bibliography{main}
}

%% file: preamble.tex
\usepackage[dvipsnames]{xcolor}
